\documentclass[conference]{IEEEtran}   % 会议模式，双盲评审无需 peerreview 选项

\usepackage{booktabs}      % 专业表格
\usepackage{multirow}      % 表格多行合并
\usepackage{amsmath}       % 数学环境
\usepackage{amssymb}       % 数学符号
\usepackage{cite}          % 引用压缩 [1]-[3]
\usepackage{url}           % URL自动断行
\usepackage{array}         % 表格列格式增强
\usepackage{dblfloatfix}   % 允许双栏浮动体放在底部
\usepackage[caption=false,font=footnotesize]{subfig} % 子图，保留IEEE标题样式
\usepackage{algorithmic}   % 算法描述（需置于 figure 环境内）

\ifCLASSINFOpdf
  \usepackage{graphicx}
  \graphicspath{{../pdf/}{../jpeg/}{../png/}}
\else
  \usepackage{graphicx}
\fi

\usepackage{cleveref}

\begin{document}

% ========== 标题 ==========
\title{MVAD: A Benchmark Dataset for Multimodal AI-Generated Video-Audio Detection}

% ========== 作者信息（双盲评审 - 匿名） ==========
% ========== 生成第一标题页 ==========
\author{
\IEEEauthorblockN{Mengxue Hu$^{1}$, Yunfeng Diao$^{*1}$, Changtao Miao$^{*2}$, Tairui Ge$^{1}$,Taize Ge$^{1}$,Zhiqing Guo$^{3}$,\\Jianshu Li$^{2}$,Zhe Li$^{2}$,
Zhongjie Ba$^{4}$, Joey Tianyi Zhou$^{5}$}
\IEEEauthorblockA{%
$^{1}$ Hefei University of Technology,
$^{2}$ Ant Group,
$^{3}$ Xinjiang University,\\
$^{4}$ Zhejiang University
$^{5}$ Agency for Science, Technology and Research\\
$^{*}$ Corresponding authors: diaoyunfeng@hfut.edu.cn; miaoct1024@gmail.com
}
}
\maketitle

% ========== NDSS 2026 版权信息（首页底部，横跨两栏） ==========
% ========== 摘要 ==========
\begin{abstract}
%AI生成视频技术的发展已超越单纯的视觉质量提升，正迅速向音视频同步生成的方向演进。然而，现有AI生成视频数据集多集中于视觉模态，即便少数包含音频信息，也主要局限于面部深度伪造场景。这一局限性无法覆盖日益多样化的多模态AI生成内容，从而严重制约了可靠检测系统的发展。为弥补这一空白，本文提出了多模态视频-音频数据集（MVAD），这是首个专门面向AI生成多模态视频-音频内容检测的综合性数据集。MVAD具有三大关键特征：（1）真正的多模态性，涵盖三种贴近真实伪造场景的模式，即伪造视频–伪造音频、伪造视频–真实音频、真实视频–伪造音频；（2）高感知质量，数据由多种最先进的生成模型生成，确保视觉与听觉质量；（3）广泛多样性，覆盖真实与动漫两种视觉风格、四类内容（人类、动物、物体、场景）以及四种视频-音频数据类型。在此基础上，本文设计了三种贴合实际应用场景的评估协议：跨生成器音视频分类任务、跨伪造模式音视频分类任务，以及用于评估对传播过程中音视频质量退化鲁棒性的退化音视频分类任务。大量实验结果表明，MVAD数据集将显著推动AI生成视频检测领域的发展。数据集已发布于 https://github.com/HuMengXue0104/MVAD。%
The development of AI-generated video technology has moved beyond mere improvement of visual quality and is rapidly evolving toward synchronized audio-video generation. However, existing AI-generated video datasets are mostly focused on the visual modality; even the few that include audio are largely restricted to facial deepfake scenarios. This limitation fails to address the increasingly diverse landscape of multimodal AI-generated content, thereby severely hindering the development of reliable detection systems. To fill this gap, this paper introduces the Multimodal Video-Audio Dataset (MVAD), the first comprehensive dataset specifically designed for detecting AI-generated multimodal video-audio content. MVAD features three key characteristics: (1) Genuine multimodality, covering three realistic forgery patterns(fake video–fake audio, fake video–real audio, and real video–fake audio); (2) High perceptual quality, achieved by employing a diverse set of state-of-the-art generative models to ensure high visual and auditory fidelity; and (3) Extensive diversity, spanning realistic and anime visual styles, four content categories (human, animal, object, scene), and four video-audio data types. On this basis, we design three evaluation protocols tailored to real-world scenarios: cross-generator audio-video classification, cross-forgery pattern audio-video classification, and degraded audio-video classification, the latter for evaluating robustness against audio-video quality degradation during propagation. Extensive experimental results demonstrate that the MVAD dataset will significantly advance the field of AI-generated video detection. The dataset is available at https://github.com/HuMengXue0104/MVAD.
\end{abstract}

% ========== 正文 ==========
\section{Introduction}
%近年来，视频生成技术的发展已不再局限于单纯提升视觉质量，在联合音视频合成方面也取得了重大突破。当前的音视频生成模型（如 Seedance 2.0、Veo 3.0、Kling 3.0）能够直接从文本或图像等极简输入出发，合成高度同步的视频内容及其匹配音频，包括连贯的环境声、逼真的音效以及与画面动作精确对齐的声音。这一能力标志着音视频合成领域迈出了重要一步。然而，随着此类技术不断降低高质量内容创作的门槛，它们同时也加剧了人们对信息安全和内容真实性的担忧，从而凸显出对 AI 生成多模态媒体的可靠检测器的迫切需求。
%现有研究在 AI 生成视频检测方面已取得实质性进展，研究者利用最先进的生成方法构建了高质量的视频数据集。然而，大多数现有数据集仅包含视觉模态，侧重于提升视频的多样性和真实感，而未生成同步的音频。少数研究已认识到这一局限并开发了多模态视频-音频数据集，但其范围仍局限于面部深度伪造。因此，缺乏一个高质量、通用的 AI 生成多模态视频-音频内容检测数据集，严重阻碍了面向真实应用场景的可靠检测器的发展。
%为填补这一空白，我们提出了 MVAD，这是首个面向多模态 AI 生成视频-音频检测的通用大规模数据集。如图所示，MVAD 具有三大关键特征：多模态性：为弥合当前多模态视频-音频生成领域的研究缺口，MVAD 模拟了三种贴近真实场景的伪造模式。高质量：MVAD 采用精心设计的构建与评估流程，融合多种最先进的视频-音频生成模型，以产生高质量的多模态内容。这种高保真性为检测任务提供了显著的判别价值。多样性：MVAD 使用了超过 20 种不同的生成器，涵盖音频生成器、视频生成器以及一体化视频-音频生成器。数据样本跨越两种视觉领域（真实风格与动漫风格），覆盖四类内容：人类、动物、物体和场景。此外，MVAD 包含四种音视频数据类型：伪造视频–伪造音频、伪造视频–真实音频、真实视频–伪造音频以及真实视频–真实音频。%
%我们设计了三个紧密贴合真实检测场景的任务：1）跨生成器音视频分类：训练好的检测器需识别来自未见生成器的视频；2）跨伪造模式音视频分类：检测器需识别具有未见伪造模式的视频；3）退化音视频分类：评估检测器在视频经历退化（如低分辨率、压缩伪影或高斯模糊）后的性能。我们在 MVAD 数据集上对多个检测器在三个任务上进行了大量实验。这些任务将极大推动通用多模态视频-音频检测器的发展，以满足迫切的社会需求。表格展示了 MVAD 与现有数据集的对比。
%本文的主要贡献总结如下：数据集构建：提出了 MVAD，这是首个用于检测 AI 生成多模态视频-音频内容的通用数据集。它涵盖了两种视觉风格、四类内容、四种音视频数据类型以及三种伪造模式下的伪造样本，填补了现有数据集在多模态、通用场景方面的空白。评估任务设计：设计了三个贴合实际应用场景的评估任务，即跨生成器音视频分类、跨伪造模式音视频分类以及退化音视频分类，为后续研究提供了系统化的评测基准。通过发布 MVAD 并建立上述三个基准任务，我们旨在为研究界提供一个开放、多元且具有挑战性的评估平台，从而促进更鲁棒、更具泛化能力的多模态伪造检测技术的发展。
%
Recently, the development of video generation technology has moved beyond merely improving visual quality, achieving significant breakthroughs in joint video-audio synthesis. Current video-audio generation models (e.g., Seedance 2.0~\cite{seedance2.0}, Veo 3.0~\cite{veo3AI2025}, Kling 3.0~\cite{klingAI2026}) are capable of directly synthesizing highly synchronized video content with matching audio from minimal inputs such as text or images. This includes coherent ambient sounds, realistic sound effects, and audio precisely aligned with on-screen actions. This capability marks an important step forward in the field of video-audio synthesis. However, as such technologies continue to lower the barrier to high-quality content creation, they simultaneously amplify concerns regarding information security and content authenticity~\cite{harm,harm1}, thereby highlighting the urgent need for reliable detectors for AI-generated multi-modal media.

\begin{figure*}[ht]
  \centering
 \includegraphics[width=1.0\textwidth, keepaspectratio]{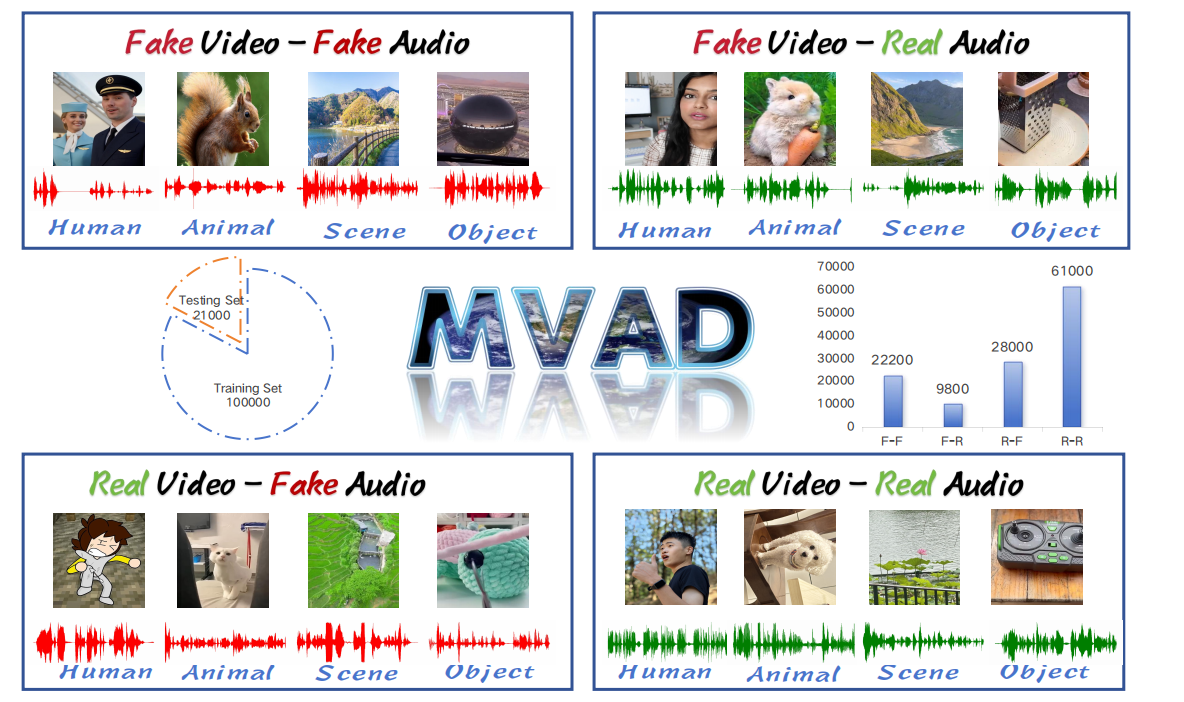}
  \caption{MVAD represents the first general-purpose dataset specifically designed for detecting AI-generated multimodal video-audio content, addressing a critical gap in current research.}
  \label{fig:statistics}
\end{figure*}

Existing research~\cite{dvf,Genvidbench,GVF,uve,demamba,genworld} has made substantial progress in AI-generated video detection, with researchers constructing high-quality video datasets using state-of-the-art generative methods. Nevertheless, most existing datasets contain only the visual modality, focusing on enhancing video diversity and realism without generating synchronized audio. A few studies~\cite{fakeav} have recognized this limitation and developed multimodal video-audio datasets, but their scope remains confined to facial deepfakes. Consequently, the lack of a high-quality, general-purpose dataset for detecting AI-generated multimodal video-audio content severely hinders the development of reliable detectors for real-world applications.

\begin{table}[t]
\caption{\textbf{Comparison with Existing AI-generated Datasets.}}
\label{datasets-comparison}
\centering
\scriptsize
\begin{tabular}{l l c c c c c}
\toprule
\textbf{Dataset} & \textbf{Publication} & \textbf{Scale} & \textbf{Multi-} & \textbf{Generation} & \textbf{Forgery} \\
& & \textbf{(k)} & \textbf{modal} & \textbf{Methods} & \textbf{Patterns} \\
\midrule
DVF\cite{dvf}         & NeurIPS 2024  & 6.7   & V & 8   & 1 \\
GenVidBench\cite{Genvidbench} & AAAI 2026  & 143  & V & 8   & 1 \\
GVF\cite{GVF}         &  arXiv 2024 & 2.8   & V & 4   & 1 \\
Uve-Bench\cite{uve}   & NeurIPS 2026 & 1.2   & V & 9   & 1 \\
GenVideo\cite{demamba}    & arXiv 2024 & 2271  & V & 20  & 1 \\
GenWorld\cite{genworld}    & arXiv 2025 & 100   & V & 9   & 1 \\
FakeAVCeleb\cite{fakeav} & NeurIPS 2021 & 20   & A\&V & 4 & 3 \\
\midrule
\textbf{MVAD}        & \textbf{-}  & 121  & A\&V & 23 & 3 \\
\bottomrule
\end{tabular}
\end{table}

To fill this gap, we propose MVAD, the first general large-scale dataset for Multimodal AI-Generated Video-Audio Detection. As presented in \Cref{fig:statistics}, MVAD features three key characteristics:

\begin{itemize}
\item \textbf{Multimodality:} To bridge the current research gap in multimodal video-audio generation, MVAD simulates three realistic forgery patterns.
\item \textbf{High Quality:} MVAD incorporates a carefully designed construction and evaluation pipeline, leveraging multiple state-of-the-art video-audio generation models to produce high-quality multimodal content. This high fidelity provides significant discriminative value for detection tasks.
\item \textbf{Diversity:} MVAD employs over twenty distinct generators, including audio generators, video generators, and integrated video-audio generators. The data samples span two visual domains (realistic and anime-style) and cover four content categories: humans, animals, objects, and scenes. Additionally, MVAD includes four audio-video data types: fake video–fake audio, fake video–real audio, real video–fake audio, and real video–real audio.
\end{itemize}

We design three tasks that closely mirror real-world detection scenarios: 
\begin{itemize}
\item\textbf{{Cross-Generator Audio-Video Classification:}} The trained detector must identify videos from unseen generators; 
\item\textbf{{Cross-Forgery Audio-Video Classification:}} The detector must identify videos with unseen forgery patterns;  \item\textbf{Degraded Audio-Video Classification:} The detector performance after video degradation (e.g., low resolution, compression artifacts, or Gaussian blur).
\end{itemize}

We conduct extensive experiments on the MVAD dataset with multiple detectors across the three tasks. These tasks will greatly advance the development of general-purpose multimodal video-audio detectors to meet pressing societal needs. \Cref{datasets-comparison} illustrates the comparison of MVAD with existing datasets.Our contributions are summarized as follows:
\begin{itemize}
    \item \textbf{Dataset Construction:} We propose MVAD, the first general-purpose dataset for detecting AI-generated multimodal video-audio content. It comprises forged video-audio samples spanning two visual styles, four content categories, four audio-video data types, and three forgery patterns, filling the gap in existing datasets for multimodal and general-domain scenarios.
    \item \textbf{Evaluation Tasks:} We design three evaluation tasks tailored to real-world applications, namely cross-generator audio-video classification, cross-forgery audio-video classification, and degraded audio-video classification, providing a systematic benchmark for future research.
\end{itemize}

With the release of MVAD and the establishment of these three benchmark tasks, we aim to provide the research community with an open, diverse, and challenging evaluation platform, thereby fostering the development of more robust and generalizable multimodal forgery detection techniques.

\section{Related Works}

\begin{table*}[t]
\caption{Statistics of Real and Generated Video-Audio Content in the MVAD Training Dataset.}
\label{dataset-table}
\vskip 0.15in
\begin{center}
\begin{small}
\begin{tabular}{lllllll}
\hline
\textbf{Video Source} & \textbf{Modality} & \textbf{Length} & \textbf{Train} & \textbf{Test} & \textbf{Count} & \textbf{Total Count}\\
\hline
UGC-Video\cite{ugc} & R-R & 10-60s & - & 1000 & \multirow{2}{*}{11000} & \multirow{4}{*}{61000}\\
HumoSet\cite{humo} & R-R & 10s & - & 10000 & & \\
\cline{1-6}
HarmonySet\cite{harmonyset} & R-R & 10-60s & 20000 & - & \multirow{2}{*}{50000} & \\
TalkVid\cite{talkvid} & R-R & 3s & 30000 & - & & \\
\hline
MSVD*\cite{msvd} & R-F & 1-10s & - & 1000*4 & 4000 & \multirow{4}{*}{28000}\\
\cline{1-6}
OpenVid-1M*\cite{openvid} & R-F & 1-10s & 2000*4 & - & \multirow{3}{*}{24000} & \\
InternVid-10M*\cite{internvid} & R-F & 1-10s & 2000*4 & - & & \\
MSR-VTT*\cite{msrvtt} & R-F & 1-10s & 2000*4 & - & & \\
\hline
Pika*\cite{pika} & F-F & 3-5s & - & 166*4 & \multirow{8}{*}{5000} & \multirow{15}{*}{22200}\\
ViduQ2\cite{vidu2025} & F-F & 2-5s & - & 1171 & & \\
ViduQ3\cite{viduQ3} & F-F & 2-5s & - & 100 & & \\
Wan2.6\cite{wan2_6} & F-F & 2-5s & - & 100 & & \\
Seedance1.5pro\cite{seedance1.5} & F-F & 5s & - & 1042 & & \\
Seedance2.0\cite{seedance2.0} & F-F & 5s & - & 270 & & \\
kling3.0\cite{klingAI2026} & F-F & 5/10s & - & 622 & & \\
Veo3\cite{veo3AI2025} & F-F & 10-60s & - & 41 & & \\
\cline{1-6}
JiMeng*\cite{jimeng} & F-F & 5-10s & 1191*4 & - & \multirow{7}{*}{17200} & \\
KlingO1\cite{klingAI2025} & F-F & 4s & 1100*4 & - & & \\
Sora2\cite{soraOpenAI2025} & F-F & 5-10s & 5000 & - & & \\
Kling2.1\cite{klingAI2025} & F-F & 5-10s & 513 & - & & \\
Kling1.6\cite{klingAI2024} & F-F & 5-10s & 324 & - & & \\
Kling2.6\cite{klingAI2025} & F-F & 5-10s & 1902 & - & & \\
kling2.5Turbo\cite{klingAI2025} & F-F & 5-10s & 297 & - & & \\
\hline
Wan2.1\cite{WanVideo2025} & F-R & 3s & - & 500 & \multirow{3}{*}{1000} & \multirow{4}{*}{9800}\\
Kling-Avatar\cite{klingAI2025} & F-R & 3s & - & 300 & & \\
HunYuan-Avatar\cite{hunyuanvideo} & F-R & 3s & - & 200 & & \\
\cline{1-6}
Humo \cite{humo} & F-R & 3s & 8800 & - & 8800 & \\
\hline
\textbf{Total Count} & - & - & \textbf{100000} & \textbf{21000} & \textbf{121000} & \textbf{121000} \\
\hline
\end{tabular}
\end{small}
\end{center}
\vskip -0.1in
\end{table*}
\subsection{Video Generation Methods}

With the rapid advancement of generative models, particularly diffusion models \cite{dm}, AI-generated video has garnered significant attention due to its broad downstream applications \cite{application,application1}. VideoPoet \cite{videopoet} employs a decoder-only transformer architecture to process multimodal inputs and generate high-quality video scenes. Both Kling \cite{klingAI2024} and Vidu \cite{vidu2025} adopt the Diffusion Transformer (DiT) architecture to produce high-quality video content across diverse scenarios.

However, the evolution of video generation technology is no longer focused solely on improving unimodal video quality; it has also achieved a major breakthrough in transitioning from unimodal video to integrated video-audio generation. Veo3 \cite{veo3AI2025} has pioneered synchronized video-audio generation, marking a significant leap forward in AI-generated video technology. Sora2 \cite{soraOpenAI2025} can natively generate audio that precisely matches visual content based on text and image prompts, enabling tasks such as dialogue generation, lip synchronization, ambient sound effects, background music, and emotional ambiance.

\subsection{AI-Generated Video Datasets}

The potential misuse of AI-generated videos for telecommunications fraud and defamatory content has raised significant concern \cite{harm}. To advance detection capabilities, numerous datasets containing both authentic and fake videos have been constructed for training and evaluation. Early AI-generated video datasets, such as DFDC \cite{DFDC}, primarily focused on deepfake detection. With the rapid development of diffusion models \cite{dm} and their variants \cite{dm1,dm2}, AI-generated video content has expanded beyond facial regions. Consequently, substantial research efforts have shifted toward building general-purpose AI-generated video datasets, including DVF \cite{dvf} and GenWorld \cite{genworld}. A representative dataset, GenVideo \cite{demamba}, incorporates 20 state-of-the-art AI-generated video models (including Pika \cite{pika} and OpenSora \cite{opensora}) with a total data volume reaching millions of samples.

However, all these datasets focus exclusively on unimodal (visual) detection, overlooking the growing prevalence of multimodal video-audio generated content. Although a few studies (e.g., FakeAVCeleb \cite{fakeav}) have recognized this gap and begun designing AI-generated multimodal video-audio datasets, their research scope remains confined to the domain of deepfakes, with AI-generated content limited to human faces and voices. This creates a substantial gap compared to real-world general video-audio content, which exhibits significantly richer semantic diversity and involves more complex scenarios.

%dataset%
\begin{figure*}[ht]
  \centering
\includegraphics[width=1.0\textwidth,trim=0 0 0 0, clip]{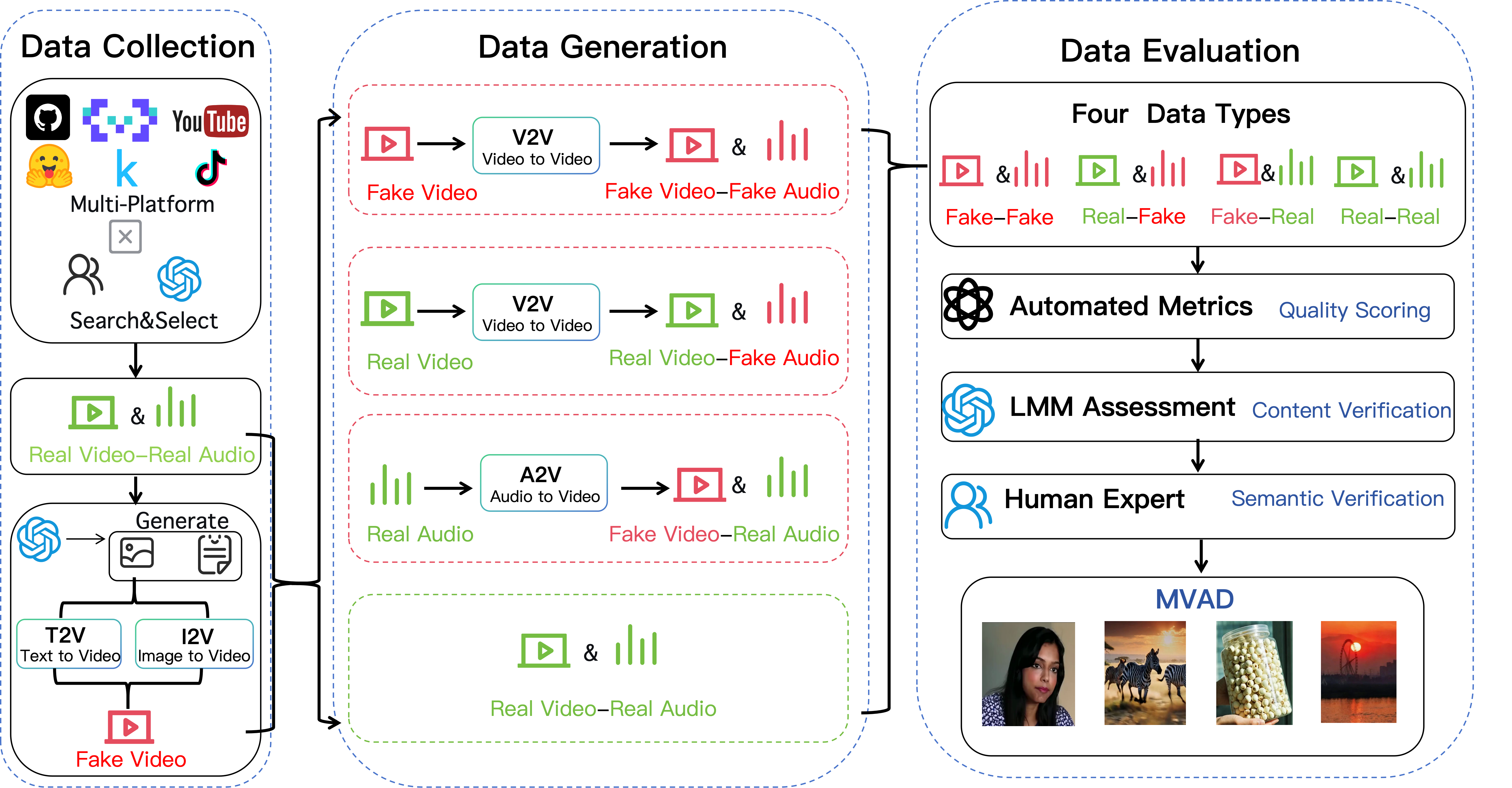}
  \caption{Construction pipeline of MVAD, comprising: data collection from open sources and self-synthesized content generation; multi-stage data generation implementing three distinct forgery patterns; and comprehensive evaluation through automated metrics, LMM assessment, and human expert verification.}
  \label{fig:framework}
\end{figure*}
\subsection{AI-Generated Video Detection}

%尽管现有研究在AI生成视频检测领域已取得一定进展，但其方法仍主要依赖于特定区域的时序伪影、数据集偏差或单一伪造类型，缺乏对通用场景下时序伪影的深入建模与可解释性分析。%
%为弥补这一不足，近期一系列工作从不同角度探索了更具泛化性和鲁棒性的检测机制：D3 [1] 引入二阶动力学建模，通过“差的差”量化真实与生成视频的时序波动差异，实现无需训练的通用检测；AVFF [2] 基于音视频多模态对应，通过自监督对比与互补掩码学习内在对齐关系，再以分类器捕捉伪造视频中的跨模态不一致；LipFD [3] 针对唇同步伪造，利用唇部运动与音频的时序不一致性，结合头部姿态辅助线索，构建双分支Transformer；AVH-Align [4] 指出主流数据集存在前导静音偏差，并转而仅在真实视频上训练音视频特征对齐网络，实现无监督的伪造检测。然而，上述方法仍受限于特定伪造类型、数据偏差或特征提取方式，难以直接推广至开放、多样的通用场景。为此，我们构建了一个面向通用场景的音视频检测数据集，并提出了相应的检测方法，旨在更全面地覆盖真实世界中的各类音视频伪造，提升检测的泛化能力与鲁棒性%

To address this limitation, recent works have explored more generalizable and robust detection mechanisms from different perspectives. D3 \cite{d3} introduces second-order dynamics modeling, quantifying the difference in temporal fluctuations between real and AI-generated videos via the “difference of differences” to achieve training-free general detection. AVFF \cite{avff} leverages audio-video multimodal correspondences, learning intrinsic alignment through self-supervised contrastive and complementary masking, and then captures cross-modal inconsistencies in forged videos with a classifier. LipFD \cite{lipfd} targets lip-sync forgery, exploiting temporal inconsistencies between lip movements and audio, augmented with head pose as auxiliary cues, within a dual-branch Transformer. AVH-Align \cite{avh} identifies the leading-silence bias prevalent in mainstream datasets and instead trains an audio-video feature alignment network solely on real videos, achieving unsupervised forgery detection. However, these methods remain constrained by specific forgery patterns, data biases, or feature extraction approaches, making them difficult to generalize directly to open and diverse real-world scenarios. To this end, we construct a general-purpose audio-video detection dataset and propose a corresponding detection method, aiming to cover various types of audio-video forgeries in real-world settings more comprehensively and to improve the generalization ability and robustness of detection.   % 请确认实际文件名（原模板为 Relatework.tex）

\section{Dataset}
%MVAD是首个面向多模态视频-音频生成检测的通用数据集，旨在反映真实世界中音视频内容的多样性。它涵盖现实与动漫两种视觉风格，以及人类、动物、物体、场景四类常见内容。数据集包含三种视频-音频伪造类型和四种模态组合（假-假、假-真、真-假、真-真），总样本量为121,000个，由20余种生成方法产生，其中伪造样本22,200个，真实样本61,000个，严格遵循1:1的伪造/真实比例。各模态组合分布为：假-假62,178、假-真28,000、真-假9,800、真-真61,000。整体上，MVAD具备三大特点：一是聚焦多模态伪造检测，填补了该领域的数据空白；二是集成了文本到视频、图像到视频、视频到视频、音频到视频等四种生成方式，覆盖丰富的伪造特征；三是贴近实际应用场景，支持对不同视觉域和内容类别的广泛评估。%
\subsection{Overview}
MVAD is the first general-purpose dataset designed for multimodal video-audio generation detection, aiming to reflect the diversity of real-world audio-visual content, as shown in \Cref{dataset-table}. It covers two visual styles (realistic and anime) and four common content categories: human, animal, object, and scene. The dataset includes three forgery patterns and four data types (fake-fake, fake-real, real-fake, real-real), with a total of 121,000 samples AI-generated by over 20 methods. Among these, there are 22,200 forged samples and 61,000 authentic samples, strictly maintaining a 1:1 forged-to-authentic ratio. The distribution across data types is as follows: fake-fake: 62,178; fake-real: 28,000; real-fake: 9,800; real-real: 61,000. Overall, MVAD has three key characteristics: (1) it focuses on multimodal forgery detection, filling a critical data gap in this field; (2) it integrates four generation approaches—text-to-video, image-to-video, video-to-video, and audio-to-video—covering a rich variety of forgery characteristics; (3) it is closely aligned with real-world application scenarios, supporting broad evaluation across different visual domains and content categories.

\subsection{Data Collection}
During the data collection phase, we acquired various input materials for constructing multimodal forgery data, including prompt combinations, fake videos, real videos, genuine audio samples, as well as real video-real audio pairs representing one of the four data types in our dataset.

\begin{table*}[t]
\caption{Video quality assessment metrics comparison across datasets}
\label{quality-metrics}
\vskip 0.15in
\begin{center}
\begin{small}
\begin{tabular}{lcccccc}
\toprule
\textbf{Dataset} & \textbf{Aesthetic} & \textbf{Background} & \textbf{Image} & \textbf{Motion} & \textbf{Subject} & \textbf{Temporal} \\
& \textbf{Quality} & \textbf{Consistency} & \textbf{Quality} & \textbf{Smoothness} & \textbf{Consistency} & \textbf{Flickering} \\
\midrule
DVF \cite{dvf}& 0.5029 & 0.9383 & 0.6103 & 0.9749 & 0.9199 & 0.9614 \\
GenVidBench \cite{Genvidbench}& 0.4625 & 0.9471 & 0.6057 & 0.9733 & 0.9490 & 0.9679 \\
GVF \cite{GVF}& 0.5142 & 0.9411 & 0.6141 & 0.9470 & 0.9200 & 0.9325 \\
Uve-Bench \cite{uve}& 0.5605 & 0.9534 & 0.5991 & 0.9782 & 0.9325 & 0.9730 \\
\midrule
\textbf{MVAD} & \textbf{0.6003} & \textbf{0.9744} & \textbf{0.7678} & \textbf{0.9793} & \textbf{0.9785} & \textbf{0.9765} \\
\bottomrule
\end{tabular}
\end{small}
\end{center}
\vskip -0.1in
\end{table*}
\begin{table*}[t]
\caption{One-to-many generalization experiment on \textbf{Real-Fake} forged modalities. We train the model on the Openvid-Hunyuan subset and test its generalization ability on other real-fake forged modality samples.}
\label{openvid_generalization}
\centering
\small
\setlength{\tabcolsep}{4pt}
\renewcommand{\arraystretch}{1.0}
\begin{tabular}{l l l l c c c c}
\toprule
\textbf{Training Subset} & \textbf{Model} & \textbf{Detection Level} & \textbf{Metric} & \textbf{MSVD-MMAudio} & \textbf{InternVid-AudioX} & \textbf{MSRVTT-FoleyCrafter} & \textbf{Avg.} \\
\midrule
\multirow{12}{*}{Openvid-Hunyuan} 
& \multirow{4}{*}{AVH-Align} & \multirow{4}{*}{Video \& Audio} & R & 0.9920 & 0.8056 & 1.0000 & 0.9325 \\
& & & F1 & 0.6627 & 0.5739 & 0.6662 & 0.6343 \\
& & & AP & 0.3177 & 0.3155 & 0.3596 & 0.3309 \\
& & & ACC & 0.4955 & 0.4024 & 0.4995 & 0.4658 \\
\cmidrule(lr){2-8}
& \multirow{4}{*}{LipFD} & \multirow{4}{*}{Video \& Audio} & R & 0.9886 & 0.9938 & 0.9972 & 0.9932 \\
& & & F1 & 0.7172 & 0.7196 & 0.7212 & 0.7193 \\
& & & AP & 0.4931 & 0.7422 & 0.5703 & 0.6019 \\
& & & ACC & 0.6105 & 0.6131 & 0.6148 & 0.6128 \\
\cmidrule(lr){2-8}
& \multirow{4}{*}{AVFF} & \multirow{4}{*}{Video \& Audio} & R & 0.6320 & 0.6420 & 0.6320 & 0.6353 \\
& & & F1 & 0.7736 & 0.7744 & 0.6358 & 0.7279 \\
& & & AP & 0.8882 & 0.9810 & 0.7214 & 0.8635 \\
& & & ACC & 0.8148 & 0.8128 & 0.6376 & 0.7551 \\
\bottomrule
\end{tabular}
\end{table*}

\noindent\textbf{Source Collection.} As illustrated in the data collection phase of \Cref{fig:framework}, we collected raw data from open datasets \cite{talkvid,msvd,openvid} and academic repositories to ensure the comprehensiveness of our dataset design. The sourced real videos and authentic video-audio content were strategically selected using Large Multimodal Models \cite{deepseek2024,gpt4o} and human expertise to cover a wide spectrum of real-world scenarios.

\noindent\textbf{Prompt Generation.} Following source collection, a comprehensive cleaning and filtering procedure was implemented with LMMs to identify and remove duplicate and low-quality samples. We subsequently extracted the first-frame images and corresponding audio tracks from the processed videos. These LMMs were then employed to generate precise and comprehensive textual prompts for AI-generated video models through analysis of the processed first-frame images and their paired audio content (where available). This process yielded two types of prompt combinations: text-image pairs and audio-text-image pairs, both of which served as the foundation for subsequent generation.

\noindent\textbf{Fake Video Generation.} In this workflow, our unimodal video data functioned as the initial material, making unimodal video generation an integral component of the data collection process. Through a synergistic combination of human expert guidance and automated LMM orchestration, we employed advanced generation methods \cite{pika,klingAI2025,jimeng} to produce diverse general AI-generated videos across various scenarios. Specifically, our approach incorporated both text-to-video and image-to-video generation techniques. Text-to-Video (T2V) generation produces content based on semantic information while preserving the model's appearance preferences, whereas Image-to-Video (I2V) generation utilizes image inputs as references to achieve superior appearance quality and semantic coherence. These complementary methods simulate authentic videos through distinct mechanisms, each offering unique analytical value for detection research.

\subsection{Data Generation}
At this stage, we complete the core component of the dataset construction pipeline—generating multimodal forged video-audio data. MVAD is built by simulating three characteristic real-world forgery patterns, employing generation models to construct diverse multimodal video-audio content.

\noindent\textbf{Fake Video-Fake Audio.} The fake video-fake audio data type comprises two distinct generation approaches: direct synthesis and indirect synthesis. The direct approach \cite{klingAI2025, soraOpenAI2025} generates synchronized video-audio content directly from text or image inputs in a unified process. In contrast, the indirect approach first generates a fake video and subsequently synthesizes a corresponding audio track based on its visual content \cite{foleycrafter, hunyuanvideo, mmaudio, audiox}. Human experts and Large Multimodal Models (LMMs) facilitate this process by employing advanced generation techniques—including video-to-AV generation for indirect synthesis and text/image-to-AV generation for direct synthesis—leveraging the text-image prompt pairs obtained during the data collection phase to produce the final fake video-fake audio samples.

\noindent\textbf{Fake Video-Real Audio.} For the generation of fake video-real audio data, both Large Multimodal Models (LMMs) and human experts employ specialized video-audio generation methods \cite{WanVideo2025, humo, klingAI2025}. By utilizing the audio-text-image prompt pairs obtained during the data collection phase, they produce convincing fake video-real audio samples through these complementary approaches.

\noindent\textbf{Real Video-Fake Audio.} The generation process for the real video-fake audio data follows a workflow analogous to the indirect generation method used for fake video-fake audio samples. Specifically, building upon the content of unimodal real videos obtained during the data collection phase, both Large Multimodal Models (LMMs) and human experts employ various advanced video-to-audio generation methods \cite{foleycrafter, hunyuanvideo, mmaudio, audiox} to complement the real video with synthetic audio.

\begin{table*}[t]
\caption{One-to-many generalization experiment on \textbf{fake-real} forged modalities. We train the model on the Humo subset and test its generalization ability on other fake-real forged modality samples.}
\label{humo}
\centering
\small
\setlength{\tabcolsep}{4pt}
\renewcommand{\arraystretch}{1.0}
\begin{tabular}{l l l l c c c c}
\toprule
\textbf{Training Subset} & \textbf{Model} & \textbf{Detection Level} & \textbf{Metric} & \textbf{Hunyuan-Avatar} & \textbf{Kling-Avatar} & \textbf{Wan-Avatar} & \textbf{Avg.} \\
\midrule
\multirow{24}{*}{Humo} 
& \multirow{4}{*}{X-CLIP} & \multirow{4}{*}{Video} & R & 0.0455 & 0.0405 & 0.0153 & 0.0238 \\
& & & F1 & 0.0870 & 0.0757 & 0.0293 & 0.0451 \\
& & & AP & 0.4613 & 0.5013 & 0.4350 & 0.4555 \\
& & & ACC & 0.5435 & 0.5058 & 0.4930 & 0.4986 \\
\cmidrule(lr){2-8}
& \multirow{4}{*}{D3} & \multirow{4}{*}{Video} & R & 0.9048 & 0.9827 & 0.9770 & 0.9758 \\
& & & F1 & 0.7037 & 0.6733 & 0.6779 & 0.6775 \\
& & & AP & 0.7097 & 0.5856 & 0.6231 & 0.6153 \\
& & & ACC & 0.6444 & 0.5231 & 0.5345 & 0.5354 \\
\cmidrule(lr){2-8}
& \multirow{4}{*}{DeMamba} & \multirow{4}{*}{Video} & R & 0.0243 & 0.1618 & 0.1658 & 0.1593 \\
& & & F1 & 0.0127 & 0.2523 & 0.2600 & 0.2485 \\
& & & AP & 0.4121 & 0.6445 & 0.6021 & 0.6075 \\
& & & ACC & 0.4222 & 0.5202 & 0.5269 & 0.5207 \\
\cmidrule(lr){2-8}
& \multirow{4}{*}{AVH-Align} & \multirow{4}{*}{Video \& Audio} & R & 1.000 & 1.000 & 1.000 & 1.000 \\
& & & F1 & 0.6471 & 0.6667 & 0.6672 & 0.6663 \\
& & & AP & 0.3068 & 0.5923 & 0.6343 & 0.5855 \\
& & & ACC & 0.4783 & 0.5000 & 0.5006 & 0.4996 \\
\cmidrule(lr){2-8}
& \multirow{4}{*}{LipFD} & \multirow{4}{*}{Video \& Audio} & R & 0.0632 & 0.2468 & 0.1008 & 0.1423 \\
& & & F1 & 0.1062 & 0.3705 & 0.1699 & 0.2266 \\
& & & AP & 0.4190 & 0.7119 & 0.5559 & 0.5967 \\
& & & ACC & 0.5302 & 0.5806 & 0.4632 & 0.5003 \\
\cmidrule(lr){2-8}
& \multirow{4}{*}{AVFF} & \multirow{4}{*}{Video \& Audio} & R & 0.1667 & 0.2486 & 0.2046 & 0.2161 \\
& & & F1 & 0.2424 & 0.3981 & 0.2597 & 0.2998 \\
& & & AP & 0.5801 & 0.9588 & 0.4305 & 0.5956 \\
& & & ACC & 0.4565 & 0.6243 & 0.4176 & 0.4800 \\
\bottomrule
\end{tabular}
\end{table*}
\begin{table*}[t]
\caption{One-to-many generalization experiment on \textbf{fake-fake} forged modalities. We train the model on the Sora2 subset and test its generalization ability on other fake-fake forged modality samples.}
\label{sora2_generalization}
\centering
\small
\setlength{\tabcolsep}{4pt}
\renewcommand{\arraystretch}{1.0}
\begin{tabular}{l l l l c c c c}
\toprule
\textbf{Training Subset} & \textbf{Model} & \textbf{Detection Level} & \textbf{Metric} & \textbf{Kling3.0} & \textbf{ViduQ2} & \textbf{Seedance2.0} & \textbf{Avg.} \\
\midrule
\multirow{24}{*}{Sora2} 
& \multirow{4}{*}{X-CLIP} & \multirow{4}{*}{Video} & R & 0.4280 & 0.3000 & 0.5490 & 0.3553 \\
& & & F1 & 0.5120 & 0.3886 & 0.4098 & 0.4432 \\
& & & AP & 0.6307 & 0.5537 & 0.5948 & 0.5927 \\
& & & ACC & 0.5916 & 0.5280 & 0.5491 & 0.5579 \\
\cmidrule(lr){2-8}
& \multirow{4}{*}{D3} & \multirow{4}{*}{Video} & R & 1.0000 & 1.0000 & 0.8220 & 0.9740 \\
& & & F1 & 0.7806 & 0.6747 & 0.6714 & 0.7206 \\
& & & AP & 0.3618 & 0.5738 & 0.6055 & 0.4800 \\
& & & ACC & 0.6402 & 0.5187 & 0.5654 & 0.4836 \\
\cmidrule(lr){2-8}
& \multirow{4}{*}{Demamba} & \multirow{4}{*}{Video} & R & 0.0440 & 0.0071 & 0.1080 & 0.0401 \\
& & & F1 & 0.0830 & 0.0140 & 0.1933 & 0.0741 \\
& & & AP & 0.6546 & 0.8233 & 0.7901 & 0.7481 \\
& & & ACC & 0.5130 & 0.5027 & 0.5514 & 0.5515 \\
\cmidrule(lr){2-8}
& \multirow{4}{*}{AVH-Align} & \multirow{4}{*}{Video \& Audio} & R & 0.8384 & 0.6787 & 0.4429 & 0.7031 \\
& & & F1 & 0.5278 & 0.4870 & 0.3626 & 0.4811 \\
& & & AP & 0.3469 & 0.3508 & 0.3178 & 0.3164 \\
& & & ACC & 0.3611 & 0.3283 & 0.2306 & 0.3246 \\
\cmidrule(lr){2-8}
& \multirow{4}{*}{LipFD} & \multirow{4}{*}{Video \& Audio} & R & 0.7394 & 0.7955 & 0.5910 & 0.7364 \\
& & & F1 & 0.6397 & 0.7151 & 0.6171 & 0.6669 \\
& & & AP & 0.5754 & 0.7047 & 0.6664 & 0.6447 \\
& & & ACC & 0.6481 & 0.6867 & 0.6376 & 0.6622 \\
\cmidrule(lr){2-8}
& \multirow{4}{*}{AVFF} & \multirow{4}{*}{Video \& Audio} & R & 0.5130 & 0.4555 & 0.4419 & 0.4816 \\
& & & F1 & 0.4767 & 0.5029 & 0.4043 & 0.4686 \\
& & & AP & 0.4168 & 0.5985 & 0.3647 & 0.4569 \\
& & & ACC & 0.4374 & 0.5490 & 0.3458 & 0.4492 \\
\bottomrule
\end{tabular}
\end{table*}

\subsection{Data Evaluation}
To ensure the high quality and practical utility of the dataset, we conduct comprehensive evaluations on samples from all four data types. The real-real video-audio data obtained during the data collection phase has been pre-filtered using LMMs based on: (1) video subject, (2) technical parameters (resolution, frame rate, etc.), and (3) content quality (visual and audio clarity). For the three forgery patterns of forged samples, we implement a three-tiered filtering pipeline with the following priority hierarchy (from highest to lowest): human expert evaluation, Large Multimodal Model (LMM) assessment, and automated quality evaluation.

\noindent\textbf{Automated Quality Metrics.} Our filtering process begins with automated quality assessment using the VBench evaluation protocol \cite{vbench}. This system evaluates the three types of video-audio forgery samples across 16 distinct dimensions, with predetermined quality thresholds established for each dimension. For instance, samples must achieve an Image Quality score above 75 to proceed to subsequent evaluation stages.

\noindent\textbf{Large Multimodal Model Assessment.} Samples that pass the automated assessment undergo evaluation by Large Multimodal Models \cite{deepseek2024}, guided by a specifically designed prompt: "Analyze the synthetic video-audio content by evaluating clarity and naturalness of both modalities, checking for noise or distortion. Assess video resolution, frame rate stability, movement naturalness, and lighting consistency. Evaluate video-audio synchronization, identify potential synthetic artifacts, and provide scores (1-10) for audio quality, video quality, and synchronization. Summarize key strengths and weaknesses, returning results in a structured format." Samples achieving minimum scores of 7 across all evaluation dimensions proceed to the final assessment stage.

\noindent\textbf{Human Expert Evaluation.} The highest-priority assessments are conducted by human experts. We established a dedicated annotation platform and recruited ten domain experts to manually evaluate each video-audio sample. Experts categorize samples into three quality tiers—low, medium, and high—retaining only those rated as medium or high quality in the final dataset.

After implementing the three-tiered filtering pipeline (prioritized as human expert evaluation, large multimodal model assessment, and automated quality evaluation), the samples in our audio-video dataset are assured of high quality. A comparison of dataset quality with other datasets is presented in \cref{quality-metrics}.
\section{MVAD Benchmark}
\subsection{Detectors}
To evaluate the MVAD dataset constructed in this work, we systematically survey current mainstream AI-generated video detection methods as well as AI-generated audio-video detection methods.

\begin{table*}[t]
\caption{Cross-modality generalization performance (selected models)}
\label{type_experiment}
\centering
\small
\setlength{\tabcolsep}{4pt}
\renewcommand{\arraystretch}{1.2}
\resizebox{\linewidth}{!}{%
\begin{tabular}{l l c c c c c c c c c c c c}
\toprule
\multirow{2}{*}{\textbf{Model}} & \multirow{2}{*}{\textbf{Metric}} &
\multicolumn{4}{c}{\textbf{Fake-Fake With Real-Fake}} &
\multicolumn{4}{c}{\textbf{Fake-Real With Real-Fake}} &
\multicolumn{4}{c}{\textbf{Fake-Fake With Fake-Real}} \\
\cmidrule(lr){3-6} \cmidrule(lr){7-10} \cmidrule(lr){11-14}
& & \textbf{Hunyuan} & \textbf{Kling} & \textbf{Wan} & \textbf{Avg.} &
\textbf{Kling} & \textbf{Vidu} & \textbf{Seedance} & \textbf{Avg.} &
\textbf{MSRVTT} & \textbf{InternVid} & \textbf{MSVD} & \textbf{Avg.} \\
& & \textbf{Avatar} & \textbf{Avatar} & \textbf{Avatar} & &
\textbf{3.0} & \textbf{Q2} & \textbf{2.0} & &
\textbf{FC} & \textbf{AX} & \textbf{MM} & \\
\midrule
\multirow{4}{*}{AVH}
& R & 0.0909 & 0.0462 & 0.0816 & 0.0716 & 0.0247 & 0.0273 & 0.0143 & 0.0237 & 0.1062 & 0.1062 & 0.1062 & 0.1062 \\
& F1& 0.1143 & 0.0650 & 0.1161 & 0.1010 & 0.0317 & 0.0361 & 0.0222 & 0.0317 & 0.1337 & 0.1337 & 0.1337 & 0.1337 \\
& AP& 0.3620 & 0.3556 & 0.3684 & 0.3609 & 0.3238 & 0.2729 & 0.3071 & 0.2755 & 0.3182 & 0.3365 & 0.3303 & 0.3283 \\
& ACC& 0.3260 & 0.3352 & 0.3780 & 0.3634 & 0.3634 & 0.3499 & 0.3788 & 0.3609 & 0.2733 & 0.3123 & 0.3123 & 0.2993 \\
\cmidrule(lr){1-14}
\multirow{4}{*}{LipFD}
& R & 0.7526 & 0.9636 & 0.7031 & 0.7816 & 0.5088 & 0.8200 & 0.3833 & 0.5850 & 0.0507 & 0.1283 & 0.0391 & 0.0727 \\
& F1& 0.6217 & 0.7919 & 0.6443 & 0.6869 & 0.4480 & 0.6859 & 0.4387 & 0.5379 & 0.0867 & 0.2057 & 0.0675 & 0.1200 \\
& AP& 0.4486 & 0.7270 & 0.6244 & 0.6480 & 0.3847 & 0.5634 & 0.4997 & 0.4979 & 0.5040 & 0.5598 & 0.4752 & 0.5130 \\
& ACC& 0.5953 & 0.7468 & 0.5766 & 0.6274 & 0.4704 & 0.6287 & 0.5153 & 0.5449 & 0.4665 & 0.5052 & 0.4607 & 0.4775 \\
\cmidrule(lr){1-14}
\multirow{4}{*}{AVFF}
& R & 0.1250 & 0.1503 & 0.1458 & 0.5850 & 0.4770 & 0.4740 & 0.5953 & 0.4966 & 0.1500 & 0.1500 & 0.1520 & 0.1507 \\
& F1& 0.2143 & 0.2613 & 0.2415 & 0.5379 & 0.4882 & 0.6431 & 0.5278 & 0.5590 & 0.2609 & 0.2595 & 0.2639 & 0.2614 \\
& AP& 0.5630 & 0.8449 & 0.6150 & 0.4979 & 0.4874 & 0.9055 & 0.4406 & 0.6685 & 0.9372 & 0.8870 & 0.8879 & 0.9040 \\
& ACC& 0.5217 & 0.5751 & 0.5428 & 0.5449 & 0.5005 & 0.7370 & 0.4650 & 0.5917 & 0.5746 & 0.5716 & 0.5756 & 0.5739 \\
\bottomrule
\end{tabular}%
}
\end{table*}
\begin{table*}[t]
\caption{Performance under different perturbations across three training forgery types}
\label{perturbation_all}
\centering
\footnotesize
\setlength{\tabcolsep}{3pt}
\renewcommand{\arraystretch}{1.0}
\resizebox{\linewidth}{!}{%
\begin{tabular}{l l *{5}{c} *{5}{c} *{5}{c}}
\toprule
\multirow{2}{*}{\textbf{Model}} & \multirow{2}{*}{\textbf{Metric}} &
\multicolumn{5}{c}{\textbf{Fake-Real}} &
\multicolumn{5}{c}{\textbf{Fake-Fake}} &
\multicolumn{5}{c}{\textbf{Real-Fake}} \\
\cmidrule(lr){3-7} \cmidrule(lr){8-12} \cmidrule(lr){13-17}
& & \textbf{Orig} & \textbf{Crop} & \textbf{Flip} & \textbf{H264} & \textbf{Noise} &
\textbf{Orig} & \textbf{Crop} & \textbf{Flip} & \textbf{H264} & \textbf{Noise} &
\textbf{Orig} & \textbf{Crop} & \textbf{Flip} & \textbf{H264} & \textbf{Noise} \\
\midrule
\multirow{4}{*}{D3}
& R & 0.9758 & 0.9725 & 0.6424 & 0.9540 & 0.9894 & 0.9740 & 0.9918 & 0.9930 & 0.9959 & 0.9926 & -- & -- & -- & -- & -- \\
& F1& 0.6775 & 0.6803 & 0.6820 & 0.6746 & 0.6725 & 0.7206 & 0.6672 & 0.6717 & 0.6687 & 0.6680 & -- & -- & -- & -- & -- \\
& AP& 0.6153 & 0.6035 & 0.6424 & 0.6092 & 0.5667 & 0.4800 & 0.5098 & 0.4831 & 0.5108 & 0.5165 & -- & -- & -- & -- & -- \\
& ACC& 0.5354 & 0.5431 & 0.6820 & 0.5403 & 0.5187 & 0.4836 & 0.5060 & 0.5143 & 0.5068 & 0.5067 & -- & -- & -- & -- & -- \\
\cmidrule(lr){1-17}
\multirow{4}{*}{Demamba}
& R & 0.1593 & 0.2400 & 0.0543 & 0.1686 & 0.0658 & 0.0401 & 0.0143 & 0.0314 & 0.0686 & 0.0286 & -- & -- & -- & -- & -- \\
& F1& 0.2485 & 0.3137 & 0.0967 & 0.2616 & 0.1215 & 0.0741 & 0.2786 & 0.0595 & 0.1263 & 0.0528 & -- & -- & -- & -- & -- \\
& AP& 0.6075 & 0.4667 & 0.5753 & 0.6089 & 0.5677 & 0.5677 & 0.4534 & 0.5892 & 0.5570 & 0.4908 & -- & -- & -- & -- & -- \\
& ACC& 0.5207 & 0.4750 & 0.4929 & 0.5243 & 0.5043 & 0.5043 & 0.5014 & 0.5029 & 0.5257 & 0.4871 & -- & -- & -- & -- & -- \\
\cmidrule(lr){1-17}
\multirow{4}{*}{AVH-Align}
& R & 1.0000 & 1.0000 & 1.0000 & 1.0000 & 1.0000 & 0.7301 & 0.6493 & 0.6295 & 0.6500 & 0.6590 & 0.9325 & 0.9920 & 0.9900 & 0.9960 & 0.9840 \\
& F1& 0.6663 & 0.6666 & 0.6663 & 0.6663 & 0.6222 & 0.4811 & 0.4605 & 0.4549 & 0.4620 & 0.4720 & 0.6343 & 0.6627 & 0.6618 & 0.6644 & 0.6626 \\
& AP& 0.5855 & 0.5834 & 0.5251 & 0.5777 & 0.5461 & 0.3164 & 0.3062 & 0.3260 & 0.3060 & 0.3070 & 0.3309 & 0.3157 & 0.3244 & 0.3154 & 0.3163 \\
& ACC& 0.4996 & 0.5000 & 0.4996 & 0.4996 & 0.4568 & 0.3246 & 0.3052 & 0.3090 & 0.3060 & 0.3140 & 0.4658 & 0.4955 & 0.4945 & 0.4975 & 0.4955 \\
\cmidrule(lr){1-17}
\multirow{4}{*}{LipFD}
& R & 0.1455 & 0.1775 & 0.1495 & 0.1487 & 0.0885 & 0.7364 & 0.7482 & 0.7765 & 0.7315 & 0.7124 & 0.9932 & 0.9888 & 0.9914 & 0.9888 & 0.9902 \\
& F1& 0.2315 & 0.2685 & 0.2381 & 0.2356 & 0.1546 & 0.6669 & 0.6790 & 0.6556 & 0.6653 & 0.6537 & 0.7193 & 0.7114 & 0.7069 & 0.7176 & 0.7082 \\
& AP& 0.6038 & 0.5867 & 0.5877 & 0.5937 & 0.6053 & 0.6447 & 0.6640 & 0.5838 & 0.6454 & 0.6387 & 0.6019 & 0.4686 & 0.4845 & 0.4913 & 0.5272 \\
& ACC& 0.5003 & 0.5299 & 0.5308 & 0.5312 & 0.5202 & 0.6622 & 0.6758 & 0.6248 & 0.6621 & 0.6528 & 0.6128 & 0.5993 & 0.5893 & 0.6113 & 0.5891 \\
\cmidrule(lr){1-17}
\multirow{4}{*}{AVFF}
& R & 0.2161 & 0.2228 & 0.2225 & 0.2208 & 0.0000 & 0.4816 & 0.5305 & 0.5358 & 0.4858 & 0.0128 & 0.6353 & 0.6540 & 0.6400 & 0.6340 & 0.0120 \\
& F1& 0.2998 & 0.3076 & 0.2968 & 0.2968 & 0.0000 & 0.4686 & 0.5053 & 0.5204 & 0.4790 & 0.0235 & 0.7279 & 0.7899 & 0.7795 & 0.7751 & 0.0237 \\
& AP& 0.5956 & 0.5979 & 0.5708 & 0.5676 & 0.6127 & 0.4569 & 0.4709 & 0.4907 & 0.4688 & 0.4158 & 0.8635 & 0.9984 & 0.9983 & 0.9982 & 0.5905 \\
& ACC& 0.4800 & 0.4841 & 0.4735 & 0.4808 & 0.4994 & 0.4492 & 0.4767 & 0.4907 & 0.4640 & 0.4709 & 0.7551 & 0.8258 & 0.8189 & 0.8158 & 0.5055 \\
\bottomrule
\end{tabular}%
}
\end{table*}

\subsection{Implementation Details}

\noindent\textbf{Backbone Model.} A backbone model can be directly used as a baseline binary classifier for real vs. fake videos without any additional design. In this paper, we select X-CLIP \cite{xclip} as a representative video backbone. Based on the Transformer architecture, it captures authenticity differences by modeling temporal dependencies among video frames.

\noindent\textbf{General Video Detector.} In the field of general forged video detection, researchers have made significant progress. Early methods mostly focused on mining forgery traces within a single modality. D3 \cite{d3} introduces second-order dynamics modeling, quantifying the difference in temporal fluctuations between real and generated videos via the “difference of differences” to achieve training-free general detection. DeMamba \cite{demamba} aims to capture spatio-temporal artifacts by analyzing inconsistencies in both spatial and temporal dimensions to identify AI-generated videos. However, the above methods are limited to the visual modality, whereas real-world audio-video forgery samples involve three forgery patterns: fake-fake (both audio and video forged), fake-real (video forged, audio real), and real-fake (video real, audio forged). Consequently, detectors relying solely on a single visual modality not only fail to cover all these forgery patterns comprehensively but are also incapable of handling real-fake samples where the visual part is entirely real. Nevertheless, the core ideas of general video detectors can still provide important inspiration for building general audio-video detectors.

\noindent\textbf{Deepfake Audio-Video Detector.} Due to the current lack of an AI-synthesized audio-video detection dataset for general scenarios, the development of general audio-video detectors is severely hindered. Existing audio-video detectors are almost all designed for deepfake face video data. Among them, AVFF \cite{avff} learns intrinsic alignment features through self-supervised contrastive learning and complementary masking to capture cross-modal inconsistencies in forged videos. LipFD \cite{lipfd} leverages temporal inconsistencies between lip movements and audio, together with head pose as auxiliary cues, within a dual-branch Transformer framework to achieve high-precision forgery detection. AVH-Align \cite{avh} exploits the leading-silence bias and trains an audio-video feature alignment network solely on real videos, thereby realizing unsupervised forgery detection without requiring forged samples. Although these methods perform well on specific deepfake datasets, they are trained in the deepfake audio-video domain and thus cannot be directly transferred to general audio-video content beyond faces. Nonetheless, the design concepts of deepfake audio-video detectors still offer valuable reference for developing detectors in general scenarios.

\subsection{Task 1: Cross-Generator Audio-Video Classification Task}
As audio-visual generation technologies continue to evolve, emerging generative models and data distributions pose severe generalization challenges for detectors. To systematically evaluate detector performance on unseen data, we propose a cross-dataset generalization evaluation task. This task encompasses two typical generalization scenarios: one-to-many generalization and many-to-many generalization.

\noindent\textbf{One-to-Many Generalization Task.} Following common practices in AI-generated video detection, we design and perform a one-to-many generalization task. Specifically, we construct a training subset by selecting samples generated by one generation method from each of the three forgery patterns. For each detection method, we train it on a single base category (i.e., the data corresponding to the selected generation method), and then evaluate its average detection performance on the test subsets corresponding to the same forgery pattern as the training subset.

\subsection{Task 2: Cross-Forgery Pattern Audio-Video Classification Task}
To investigate the generalization ability of detection methods to unknown forged modality samples, we design a cross-forgery-pattern classification task. Specifically, based on the three forgery patterns, we construct four training-testing combinations: fake-fake with fake-real, fake-fake with real-fake, and fake-real with real-fake. For each combination, we select generation methods from two different forgery patterns to build a training subset, aiming to enable the model to learn forged features from both video and audio modalities from the training data of each combination. For each detection method, we train it separately on each training subset, and then evaluate its average detection performance on the test subset corresponding to the forgery pattern that is not included in the training subset. It should be noted that for detection methods using only a single video modality, we do not perform the cross-forgery-pattern classification task. This is because the video part in the real-fake pattern is real, and such methods cannot recognize its forged features, thus failing to identify this type of forged sample.

\subsection{Task 3: Degraded Audio-Video Classification Task}
During the transmission of audio-visual content, various types of data degradation often occur, such as compression due to network transmission, platform-based watermarking, codec format conversion, and environmental noise interference. These perturbations pose challenges to the detector’s judgment. Therefore, the robustness of the detector to perturbations is equally critical in practical detection scenarios. To this end, we select four representative perturbation types—H.264 compression, Gaussian noise, cropping, and flipping—and systematically examine their impact on detector performance. The specific parameters and implementation details of each perturbation are shown in the table. Furthermore, we evaluate the comprehensive effect of these perturbations on detection performance using models trained on the one-to-many generalization classification task based on the three forgery patterns, employing various methods.
\section{Experiment}

\subsection{Implementation details}
To comprehensively evaluate the performance of various detectors, we divide the dataset into two distinct parts: the basic training set $D_{\text{train}}$ and the out-of-domain test set $D_{\text{test}}$. Both $D_{\text{train}}$ and $D_{\text{test}}$ contain audio-video samples of three forgery patterns generated by different methods, as well as real audio-video samples from diverse sources. Specifically, $D_{\text{train}}$ comprises 50,000 real audio-video samples and 50,000 forged audio-video samples (covering all three forgery patterns). $D_{\text{test}}$ comprises 11,000 real audio-video samples and 10,000 forged audio-video samples.

Consistent with the methodologies employed in prior studies, our evaluation framework primarily reports accuracy (ACC) to measure the effectiveness of the detectors, with AP, F1, and recall (R) as supplementary evaluation metrics. The accuracy calculation is based on a threshold value of 0.5. When computing accuracy, we use the test data from the generative model itself to assess the model's ability to distinguish content generated by that specific model. For AP, F1, and recall, we incorporate the real video test set to ensure a more comprehensive and accurate evaluation.All experiments were conducted on a system equipped with four NVIDIA A100-SXM4-80GB GPUs and an Intel(R) Xeon(R) Platinum 8368 CPU @ 2.40GHz.

\subsection{Generalization across Unseen Generators}
We evaluate how well detectors trained on one generator generalize to other unseen generators (one‑to‑many). \Cref{openvid_generalization,humo,sora2_generalization}present results for real‑fake, fake‑real, and fake‑fake modalities, respectively.

\textbf{Real-Fake.} Only multimodal methods are applicable because the video is real.As shown in \Cref{openvid_generalization} AVFF achieves the best average ACC (75.5\%) and AP (86.4\%), substantially outperforming LipFD and AVH‑Align. LipFD achieves moderate performance ($ACC \approx 61\%$), while AVH‑Align performs poorly ($ACC \approx 50\%$). This indicates that AVFF's cross‑modal alignment features are more effective for detecting real‑fake forgeries.

\textbf{Fake-Real.}In \Cref{humo}, D3 achieves very high recall (97.6\% avg.) but moderate F1 (67.8\%). LipFD obtains the highest average F1 (71.9\%) and ACC (61.3\%). AVH‑Align shows perfect recall but very low precision ($AP \approx 58.6\%$), indicating high false positives. DeMamba performs poorly across most unseen generators.

\textbf{Fake-Fake.}As shown in \Cref{sora2_generalization},LipFD again achieves the best average F1 (66.7\%) and ACC (66.2\%). D3 performs well on Kling3.0 (100\% recall) but drops on Seedance2.0 (82.2\%), showing generator‑specific overfitting. DeMamba collapses on most unseen generators (F1<10\%). Multimodal methods (AVFF, AVH‑Align) also struggle, with ACC near 50\% on some targets. These results highlight that fake‑fake detection remains an open challenge.

\subsection{Generalization across Unseen Forgery Patterns.}
We test cross‑modality generalization (e.g., trained on fake‑fake, tested on real‑fake). \Cref{type_experiment} reports three combinations. Only multimodal methods are evaluated.

All methods show significant performance drops when the forgery patterns changes. For example, when training on fake‑fake and testing on real‑fake, the highest average F1 is only 68.7\% (LipFD), and ACC falls below 63\%. AVFF’s AP drops from over 90\% in within‑modality tests to 64.8\%. Similar degradation is observed for other combinations. These results indicate that current multimodal detectors learn forgery patterns that are tightly coupled with the specific modality combination, and they fail to generalize to unseen forgery patterns.
\subsection{Robustness to Degraded Inputs.}We evaluate robustness under four perturbations: crop, flip, H.264 compression, and Gaussian noise (\Cref{perturbation_all}). For fake‑real forgeries, D3 and DeMamba show relatively stable recall and F1 under crop and flip, while AVH‑Align and LipFD maintain almost unchanged metrics. For fake‑fake forgeries, LipFD demonstrates the best robustness (F1 stays between 65.3\% and 67.9\%). AVFF is more sensitive to H.264 and noise (F1 drops to 2.35\% under noise). For real‑fake forgeries, AVFF is the most robust, achieving high ACC (75.5–82.6\%) and AP (over 99\% for crop and flip). Overall, no single detector is robust across all forgery patterns and distortions. H.264 compression is the most harmful perturbation.

% ========== 致谢 ==========
\section{Conclusion}
We propose MVAD, the first general dataset specifically designed for detecting multimodal AI‑generated video‑audio content. MVAD features three key characteristics: genuine multimodality with three realistic forgery patterns (fake‑fake, fake‑real, real‑fake), high perceptual quality achieved by over 20 state‑of‑the‑art generative models, and extensive diversity covering two visual styles (realistic and anime), four content categories (human, animal, object, scene), and four video‑audio data types. Based on MVAD, we design three evaluation protocols tailored to real‑world scenarios: cross‑generator audio‑video classification, cross‑forgery patterns audio‑video classification, and degraded audio‑video classification. Through extensive experiments on six representative detectors, we validate the challenge and necessity of MVAD as a benchmark for general‑domain multimodal forgery detection.

Experimental results reveal that: single‑modality detectors completely fail on real‑fake forgeries, confirming the necessity of multimodal detection; cross‑generator generalization varies greatly across methods, with LipFD being the most consistent yet achieving only 66.7\% best F1 , far from practical deployment; cross‑forgery patterns generalization remains an open problem, as all methods suffer severe performance degradation when the forgery patterns changes, indicating that they learn modality‑combination‑specific superficial correlations rather than universal forgery representations; robustness is inconsistent, with H.264 compression being the most harmful perturbation. The main limitation is that the number of generation methods in the current dataset still needs to be increased, especially for the Fake-Real modality. Future work will include expanding with more generation methods and content categories. We hope MVAD will drive substantial progress in the field of general‑purpose multimodal AI‑generated video‑audio detection.
\section{Broader Impact}
Our primary goal is to fill the gap in current research on audio-video detection in general scenarios, enhance the detection capability of AI-generated audio-video content, and thereby protect society from deepfake fraud, disinformation, and political manipulation. To this end, we have constructed the first audio-video forgery dataset for general scenarios—MVAD—and designed three rigorous evaluation protocols, providing researchers with a more comprehensive benchmarking standard to facilitate the development of more generalized and robust detection models. The real audio-video samples in MVAD are sourced from publicly available datasets and are used within the scope of their licenses. The source data for generated videos are also derived from public datasets or open-source forums, with some sampled data taken from publicly available demos on the official websites of commercial audio-video generation models.

% ========== 参考文献 ==========
\bibliographystyle{IEEEtran}
\bibliography{main}

\end{document}